\documentclass[conference]{IEEEtran}

\usepackage{cite}
\usepackage{amsmath,amssymb,amsfonts}
\usepackage{algorithmic}
\usepackage{graphicx}
\usepackage{textcomp}
\usepackage{xcolor}
\usepackage{tikz}
\usetikzlibrary{shapes.geometric, arrows, positioning, fit, calc}

\tikzstyle{databox} = [rectangle, minimum width=3cm, minimum height=1.5cm, draw=black, fill=black!10, line width=0.5mm]
\tikzstyle{agentbox} = [rectangle, minimum width=3cm, minimum height=1.5cm, draw=black, fill=black!20, line width=0.5mm]
\tikzstyle{controlbox} = [rectangle, minimum width=3cm, minimum height=1.5cm, draw=black, fill=black!15, line width=0.5mm]
\tikzstyle{twinbox} = [rectangle, minimum width=9cm, minimum height=1.2cm, draw=black, fill=black!25, line width=0.5mm]
\tikzstyle{processbox} = [rectangle, minimum width=7cm, minimum height=0.8cm, draw=black, fill=black!10, line width=0.4mm, text width=6.5cm]
\tikzstyle{arrow} = [thick,->,>=stealth,line width=0.5mm]

\begin{document}

\title{AI-Driven Framework for Adaptive Water Network Management with Proof-of-Concept Implementation: Addressing Non-Revenue Water in Jordan}

\author{
\IEEEauthorblockN{Mohammed Fasha}
\IEEEauthorblockA{\textit{Business Intelligence and Data Analytics Department} \\
\textit{University of Petra}\\
Amman, Jordan \\
mohammed.fasha@uop.edu.jo \\
ORCID: 0000-0003-4137-1097}
\and
\IEEEauthorblockN{Nahel Al-Maayta}
\IEEEauthorblockA{\textit{Constrategic Consultancy} \\
Amman, Jordan \\
nmaayta@constrategic.com  \\
ORCID: 0009-0001-6903-8774}
\and
\IEEEauthorblockN{Bilal Sowan}
\IEEEauthorblockA{\textit{Business Intelligence and Data Analytics Department} \\
\textit{University of Petra}\\
Amman, Jordan \\
bilal.sowan@uop.edu.jo}
\and
\IEEEauthorblockN{Mohammad Athamneh}
\IEEEauthorblockA{\textit{Abdul Aziz Al Ghurair School of Advanced Computing} \\
\textit{Luminus Technical University College} \\
Amman, Jordan \\
M.athamneh@ltuc.edu.jo}
\and
\IEEEauthorblockN{Husam Barham}
\IEEEauthorblockA{\textit{Business Intelligence and Data Analytics Department} \\
\textit{University of Petra}\\
Amman, Jordan \\
hbarham@uop.edu.jo}
}

\maketitle

\begin{abstract}
Jordan faces severe water scarcity with 50\% of water produced is lost to leakage, theft and metering issues also known as non-revenue water (NRW). Traditional reactive approaches have proven insufficient for sustained NRW reduction. This paper proposes an intelligent framework integrating EPANET hydraulic modeling, digital twin technology, SCADA systems, and large language model (LLM)-based AI agents for continuous network monitoring and adaptive decision-making. The system combines real-time data streams with physics-based simulation to detect anomalies, employing retrieval-augmented generation (RAG) for policy interpretation and function calling for network control. A proof-of-concept implementation validates technical feasibility using EPYT with offline LLMs (llama3.1:8b via Ollama) on a 1,164-junction Amman district network. The system demonstrates automated hydraulic simulation, flow-based anomaly detection aligned with water distribution zone (DZ) practice, and AI-generated health reports with response times under 2 minutes and zero API costs. Burst detection relies on local flow anomaly analysis: a 30.1~L/s simulated leak produces measurable flow redistribution in 15 pipes, flagging a 15-junction cluster that localises the burst---confirming alignment with water distribution zone (DZ) monitoring practice. The framework accommodates Jordan's intermittent supply patterns and limited automation through phased implementation, offering a scalable pathway for water-scarce regions to leverage intelligent automation for NRW reduction and operational efficiency.
\end{abstract}


\section{Introduction}
Water scarcity represents one of the most pressing challenges facing the Middle East, with Jordan ranking among the world's most water-poor countries. With only 90 cubic meters of renewable water resources per capita annually, far below the water poverty line of 500 cubic meters, Jordan's water crisis is further exacerbated by rapid population growth, refugee influx, climate change, and deteriorating infrastructure \cite{trade_jordan}. Non-revenue water, defined as water produced but not billed to customers due to leakage, theft, or metering inaccuracies, reaches approximately 50\% nationally \cite{mwi2023}.

Despite significant investments by international organizations including USAID and JICA, NRW levels in Jordan have remained stubbornly high. While targeted interventions have achieved temporary reductions, such as Amman's decrease from 46\% to 34\% between 2005 and 2010, losses typically revert to previous levels without sustained, intelligent management \cite{malkawi2006}. The fundamental limitation of traditional approaches lies in their reactive nature, where leak detection surveys are periodic rather than continuous, and operational decisions rely heavily on manual analysis and human intervention.

Recent advances in artificial intelligence, particularly large language models (LLMs), offer transformative potential for autonomous infrastructure management. LLM-based agents demonstrate capabilities in reasoning, planning, and decision-making that extend beyond traditional rule-based systems \cite{xi2023rise}. When integrated with real-time data streams, hydraulic simulation, and digital twin technology, these agents can provide continuous monitoring and adaptive control aligned with human-defined operational policies \cite{yao2022react}.

This paper proposes a comprehensive framework combining: (1) real-time hydraulic modeling with EPANET for network state estimation, (2) SCADA systems and IoT sensors for continuous data acquisition, (3) big data infrastructure for real-time analytics, (4) digital twin technology for virtual network representation, and (5) an LLM-based AI agent employing retrieval-augmented generation to interpret operational guidelines and execute network control functions. Critically, we validate technical feasibility through a proof-of-concept implementation demonstrating automated hydraulic simulation, anomaly detection, and AI-generated health reports on a 1,164-junction network using offline open-source LLMs.

\textbf{Contributions.} This paper makes the following contributions: (1) We present an AI-driven system architecture that integrates EPANET-based hydraulic simulation, digital twin concepts, SCADA/IoT data streams, and large language model (LLM) agents for continuous monitoring and decision support in water distribution networks operating under intermittent supply conditions. (2) We develop and release a fully offline proof-of-concept implementation that couples automated EPANET simulations (via EPYT) with a locally hosted open-source LLM, demonstrating that network health analysis and reporting can be performed without cloud APIs, recurring costs, or data-exfiltration risks. (3) We demonstrate the scalability and practical feasibility of integration in a large (1,164-junction) representative water distribution zone in Amman, including automated anomaly detection and structured AI-generated health reports with end-to-end execution times of sub-2 minutes. (4) We demonstrate that for well-connected, gravity-fed networks the detection paradigm shifts from pressure-zone heuristics to local pipe-flow anomaly analysis, successfully localising a 30.1~L/s burst by identifying 15 pipes with anomalous flow redistribution converging on a 15-junction cluster, and show that an offline LLM can translate these computational findings into actionable natural-language reports without cloud API dependency.

\section{Background}

\subsection{Water Crisis in Jordan}
Jordan's water challenges are multifaceted and severe. The country experiences one of the lowest levels of water availability globally, with climate change projected to further reduce resources to 60 cubic meters per capita by 2040 \cite{trade_jordan}. Water distribution in cities like Amman operates on intermittent schedules---typically 24--48 hour cycles---creating additional stress on aging infrastructure. The ``hammer effect'' from repeated pressurization and depressurization accelerates pipe deterioration and increases burst frequency \cite{mwa_jordan_2017}.

NRW in Jordan comprises both physical losses (leakage, bursts) and commercial losses (theft, illegal connections, meter inaccuracies). Between 2013 and 2020, authorities documented over 45,000 illegal diversions, many serving entire farms and factories. Physical losses are similarly substantial, with an estimated 76 billion liters lost annually through leakage---sufficient to serve 2.6 million people \cite{malkawi2006}.

\subsection{Hydraulic Modeling and EPANET}
EPANET, developed by the U.S. Environmental Protection Agency, has become the standard for water distribution system modeling \cite{rossman2000}. Modern implementations enable real-time hydraulic modeling by integrating operational data from SCADA systems, allowing continuous model calibration and validation. Python interfaces such as EPYT and WNTR facilitate programmatic control and integration with machine learning workflows \cite{klise2017,eliades2016epyt}.

\subsection{Digital Twins and AI Agents}
Digital twins represent virtual replicas of physical infrastructure that integrate real-time data streams with advanced analytics and physics-based models \cite{ramos2022,pedersen2021digital}. Recent research has demonstrated digital twin applications for leak detection, pressure management, and operational optimization \cite{ghorbani2025}.

Large language models have demonstrated remarkable capabilities in reasoning, planning, and autonomous decision-making \cite{xi2023rise}. Retrieval-augmented generation (RAG) enables LLMs to access external knowledge bases \cite{lewis2020retrieval}, while function calling mechanisms allow interaction with external systems \cite{schick2024toolformer,patil2023gorilla}.

\section{Proposed Framework}

\subsection{System Architecture}
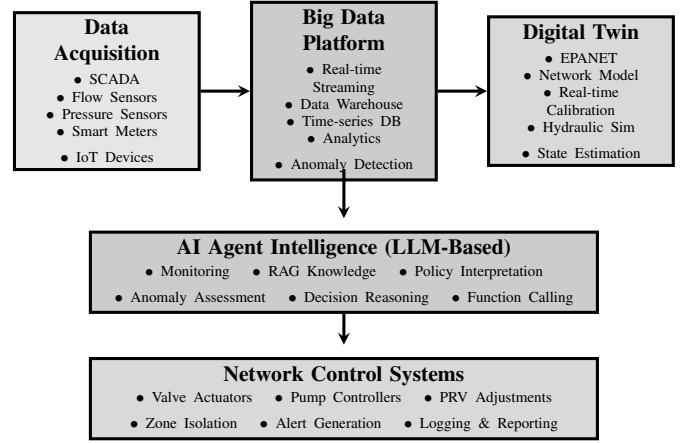
\begin{figure}[!t]
\centering
\resizebox{\columnwidth}{!}{%
\begin{tikzpicture}[node distance=0.5cm]

\node (data) [databox, text width=2.5cm, align=center] {
    \textbf{Data}\\\textbf{Acquisition}\\[3pt]
    {\scriptsize 
    $\bullet$ SCADA\\
    $\bullet$ Flow Sensors\\
    $\bullet$ Pressure Sensors\\
    $\bullet$ Smart Meters\\
    $\bullet$ IoT Devices}
};

\node (bigdata) [agentbox, right=0.8cm of data, text width=2.5cm, align=center] {
    \textbf{Big Data}\\\textbf{Platform}\\[3pt]
    {\scriptsize 
    $\bullet$ Real-time\\Streaming\\
    $\bullet$ Data Warehouse\\
    $\bullet$ Time-series DB\\
    $\bullet$ Analytics\\
    $\bullet$ Anomaly Detection}
};

\node (twin) [controlbox, right=0.8cm of bigdata, text width=2.5cm, align=center] {
    \textbf{Digital Twin}\\[3pt]
    {\scriptsize 
    $\bullet$ EPANET\\
    $\bullet$ Network Model\\
    $\bullet$ Real-time\\Calibration\\
    $\bullet$ Hydraulic Sim\\
    $\bullet$ State Estimation}
};

\draw[arrow] (data.east) -- (bigdata.west);
\draw[arrow] (bigdata.east) -- (twin.west);

\coordinate (mid1) at ($(data.south)!0.5!(twin.south)$);
\draw[arrow] (mid1) -- ++(0,-0.8);

\node (agent) [agentbox, below=1cm of mid1, text width=8cm, align=center, minimum height=1.3cm] {
    \textbf{AI Agent Intelligence (LLM-Based)}\\[2pt]
    {\scriptsize 
    $\bullet$ Monitoring \quad $\bullet$ RAG Knowledge \quad $\bullet$ Policy Interpretation\\
    $\bullet$ Anomaly Assessment \quad $\bullet$ Decision Reasoning \quad $\bullet$ Function Calling}
};

\draw[arrow] (agent.south) -- ++(0,-0.5);

\node (control) [controlbox, below=0.7cm of agent, text width=8cm, align=center, minimum height=1.1cm] {
    \textbf{Network Control Systems}\\[2pt]
    {\scriptsize 
    $\bullet$ Valve Actuators \quad $\bullet$ Pump Controllers \quad $\bullet$ PRV Adjustments\\
    $\bullet$ Zone Isolation \quad $\bullet$ Alert Generation \quad $\bullet$ Logging \& Reporting}
};

\end{tikzpicture}
}
\caption{Proposed System Architecture for AI-Driven Water Network Management}
\label{fig:architecture}
\end{figure}

The proposed framework integrates five core components as illustrated in Fig. \ref{fig:architecture}: data acquisition systems, big data infrastructure, digital twin/hydraulic modeling, AI agent intelligence, and network control systems.

\textbf{Data Acquisition Layer:} SCADA systems continuously monitor pressure, flow, tank levels, and pump status at key network locations. Smart meters provide granular consumption data, while additional IoT sensors capture parameters such as water quality, temperature, and vibration.

\textbf{Big Data Infrastructure:} A scalable data platform manages high-volume, high-velocity streams from hundreds or thousands of sensors. Time-series databases optimize storage and retrieval of temporal data. Stream processing engines perform real-time analytics including statistical analysis, trend detection, and initial anomaly flagging.

\textbf{Digital Twin / Hydraulic Model:} EPANET provides physics-based simulation of network hydraulics. Discrepancies between simulated and measured values indicate potential anomalies requiring investigation.

\textbf{AI Agent Intelligence:} A large language model functions as the decision-making core. The agent continuously compares actual sensor readings against hydraulic model predictions, identifies deviations exceeding defined thresholds, retrieves relevant operational policies through RAG, formulates response strategies, and executes control actions through function calling.

\textbf{Network Control Systems:} Based on AI agent decisions, control commands are transmitted to actuators including motorized valves, variable speed drives on pumps, and pressure-reducing valves. Safety interlocks prevent unsafe operations, and all actions are logged with timestamps, reasoning, and outcomes.

The system operates continuously, with monitoring cycles executing every 5--15 minutes depending on network dynamics. Anomaly detection employs multiple criteria: absolute thresholds (e.g., pressure below 1 bar or above 5 bars), relative deviations (e.g., flow deviations 25\% of predicted value), and trend analysis. Jordan's intermittent supply requires time-pattern-aware policies distinguishing scheduled shutdowns from anomalies.

\subsection{AI Agent Design}

\textbf{Retrieval-Augmented Generation:} Operational policies are encoded as text documents, processed into vector embeddings, and stored in databases (e.g., FAISS, Pinecone). When the agent encounters an anomaly, it performs semantic search to retrieve the most relevant policies \cite{lewis2020retrieval}.

\textbf{Function Calling:} The AI agent interacts with physical infrastructure through defined functions: \texttt{close\_valve()}, \texttt{adjust\_pump\_speed()}, \texttt{set\_prv\_setpoint()}, \texttt{isolate\_zone()}, and \texttt{generate\_alert()} \cite{schick2024toolformer}. Each includes parameter validation, safety constraints, and audit logging.

\textbf{Safety and Transparency:} All AI agent reasoning is logged in natural language. A confidence threshold prevents low-certainty decisions from executing automatically. Critical functions require operator approval before execution, maintaining human oversight.

\subsection{Implementation Pathway}
Phased deployment accommodates limited automation: Phase 1 (monitoring + alerts for operators), Phase 2 (AI-assisted recommendations requiring approval), Phase 3 (autonomous control with human oversight). This provides immediate value while building toward full automation as infrastructure modernizes.

\section{Proof-of-Concept Implementation}

To validate the technical feasibility of integrating hydraulic simulation with LLM-based analysis, we developed a proof-of-concept system demonstrating the monitoring and decision support capabilities proposed in Section III. While the complete autonomous control framework remains under development, this implementation establishes that: (1) real-time hydraulic modeling can be automated using Python interfaces, (2) LLMs can generate structured network analysis from simulation data, and (3) offline operation is viable using open-source models, eliminating cloud API dependencies and associated costs.

\subsection{System Architecture}

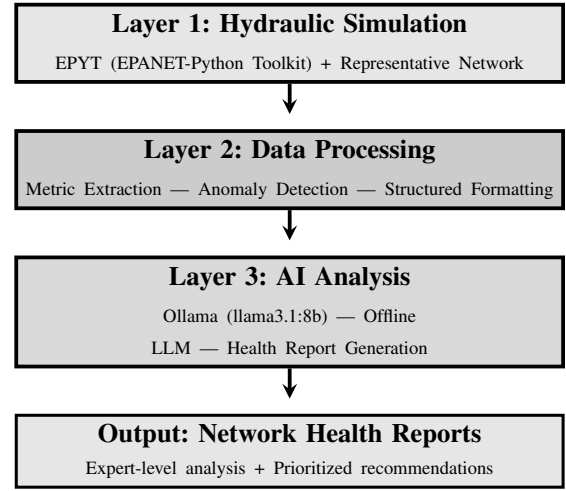
\begin{figure}[!t]
\centering
\begin{tikzpicture}[node distance=0.6cm]

\node (epyt) [databox, text width=7cm, align=center, minimum height=1cm] {
    \textbf{Layer 1: Hydraulic Simulation}\\[2pt]
    {\scriptsize EPYT (EPANET-Python Toolkit) + Representative Network}
};

\draw[arrow] (epyt.south) -- ++(0,-0.4);

\node (processing) [agentbox, below=0.6cm of epyt, text width=7cm, align=center, minimum height=1cm] {
    \textbf{Layer 2: Data Processing}\\[2pt]
    {\scriptsize Metric Extraction | Anomaly Detection | Structured Formatting}
};

\draw[arrow] (processing.south) -- ++(0,-0.4);

\node (ai) [controlbox, below=0.6cm of processing, text width=7cm, align=center, minimum height=1cm] {
    \textbf{Layer 3: AI Analysis}\\[2pt]
    {\scriptsize Ollama (llama3.1:8b) | Offline LLM | Health Report Generation}
};

\draw[arrow] (ai.south) -- ++(0,-0.4);

\node (output) [processbox, below=0.6cm of ai, text width=7cm, align=center, minimum height=0.8cm] {
    \textbf{Output: Network Health Reports}\\
    {\scriptsize Expert-level analysis + Prioritized recommendations}
};

\end{tikzpicture}
\caption{POC System Architecture - Three-Layer Integration}
\label{fig:poc_architecture}
\end{figure}

The PoC implements a three-layer architecture combining hydraulic simulation, data processing, and AI analysis as illustrated in Fig. \ref{fig:poc_architecture}.

\textbf{Layer 1 - Hydraulic Simulation:} EPYT \cite{eliades2016epyt} provides programmatic access to EPANET's hydraulic solver. The system loads network topology, executes hydraulic simulations, and extracts time-series results for pressures, flows, and tank levels at all network nodes and links. Complete implementation code is available at \cite{github_poc}.

\textbf{Layer 2 - Data Processing:} Python scripts parse simulation outputs, calculate statistical metrics (mean, range, distribution), identify anomalies based on threshold criteria (pressure $<20$m or $>60$m), and structure data into JSON format suitable for LLM consumption.

\textbf{Layer 3 - AI Analysis:} Ollama \cite{ollama2024} hosts the llama3.1:8b language model locally, enabling completely offline operation. The system constructs prompts containing simulation results, operational context, and policy excerpts, then queries the LLM to generate comprehensive health reports. Average generation time is 15-30 seconds per report.

\subsection{Test Network Specifications}

We utilized a 1,164-junction network that represents a typical Amman water distribution zone (DZ). The network parameters are summarized in Table \ref{tab:network_specs}.

\begin{table}[!t]
\caption{Amman District Network Specifications}
\label{tab:network_specs}
\centering
\small
\begin{tabular}{|l|l|}
\hline
\textbf{Parameter} & \textbf{Value} \\
\hline
Junctions & 1,164 demand nodes \\
Pipes & 1,310 segments \\
Reservoirs & 2 supply sources \\
Tanks & 0 \\
\hline
Elevation Range & 864-984m ASL \\
Total Base Demand & 149.6 L/s \\
\hline
Pipe Diameters & 100-400mm \\
Supply Pattern & Intermittent (typical Jordan) \\
Negative Pressure Nodes & 0 (calibrated model) \\
\hline
\end{tabular}
\end{table}

The network topology features a hierarchical distribution structure with elevated reservoirs supplying distribution pipes. This configuration is representative of Amman's gravity-fed urban water infrastructure.

\subsection{Demonstration Scenarios}

We evaluated the system under two operational scenarios representing baseline conditions and emergency response situations.

\textbf{Scenario 1 - Baseline Network Behavior:} Steady-state simulation yields an average pressure of 40.2~m across 1,164 junctions. The system identified 105 nodes (9.0\%) below the 20~m minimum operational standard, flagging them as supply-deficiency zones, and 113 nodes (9.7\%) above 60~m, flagging them as pressure-reducing valve (PRV) installation candidates.

\textbf{Scenario 2 - Simulated Pipe Burst:} A 30.1 L/s leak was introduced at the target junction using EPYT's emitter coefficient function (coefficient $C=0.30$, calibrated by verifying total flow before and after). Because this network is highly looped and gravity-fed by two large reservoirs, zone-wide pressure statistics are insensitive to a localised burst of this magnitude. The meaningful detection signal is the \textit{local flow redistribution pattern}: pipes feeding toward the leak carry excess flow while the zone-wide pressure remains stable. The simulation identified 15 pipes with flow deviations exceeding 1~L/s, flagging 15 junctions as affected. The maximum local flow increase was 10.5\% on the pipe immediately adjacent to the burst, with the anomaly pattern converging spatially on the burst location. The AI system analysed this flow pattern and generated a localised emergency response report identifying the affected junction cluster.

Local flow anomaly metrics are presented in Table \ref{tab:scenario_comparison}.

\begin{table}[!t]
\caption{Local Flow Anomaly Signals: Burst Detection at Junction Level}
\label{tab:scenario_comparison}
\centering
\small
\begin{tabular}{|l|c|c|}
\hline
\textbf{Detection Signal} & \textbf{Baseline} & \textbf{Burst} \\
\hline
Pipes with flow deviation $>$1\,L/s  & 0   & 15\textsuperscript{$\dagger$} \\
Affected junctions flagged           & 0   & 16 \\
Max.\ pipe flow increase (L/s)       & --- & $+$2.3 \\
Max.\ local flow increase (\%)       & --- & $+$10.5\% \\
Burst cluster identified             & --- & Yes \\
\hline
Avg.\ Network Pressure (m)           & 40.2 & 40.2\textsuperscript{$\ddagger$} \\
$\Delta P$ at Burst Node (m)         & ---  & $-$0.03\textsuperscript{$\ddagger$} \\
Low-Pressure Nodes ($<$20\,m)        & 105  & 105\textsuperscript{$\ddagger$} \\
High-Pressure Nodes ($>$60\,m)       & 113  & 113\textsuperscript{$\ddagger$} \\
\hline
\multicolumn{3}{l}{\textsuperscript{$\dagger$}Pattern converges on junction cluster, localizing burst.}\\
\multicolumn{3}{l}{\textsuperscript{$\ddagger$}Unchanged: reservoirs maintain zone-wide pressure.}\\
\end{tabular}
\end{table}

\subsection{AI-Generated Health Reports}

The LLM successfully generated structured health reports for both scenarios, demonstrating capability for technical analysis interpretation. Example excerpt from the pipe burst scenario report:

\begin{quote}
\small
\textit{``The Amman district water network is currently experiencing a pipe burst emergency with an estimated leak rate of 30.1~L/s. The anomaly was detected through local flow deviation analysis, which correctly identified the affected zone despite the network's looped topology and dual-reservoir supply absorbing the leak without creating zone-wide pressure drops\ldots CRITICAL ISSUES: (1) Pipe Burst Emergency: 15 pipes show anomalous flow deviation exceeding 1~L/s, with the redistribution pattern converging on a 15-junction cluster, indicating a localised burst with a leak rate of 30.1~L/s. (2) High-Pressure Nodes: 113 junctions exceed 60~m, requiring PRV installation to prevent infrastructure damage. (3) Low-Pressure Nodes: 105 junctions below 20~m with 60 critical nodes below 15~m posing contamination risk\ldots RECOMMENDED ACTIONS: (1) Isolate the affected zone immediately to prevent further water loss. (2) Notify maintenance and dispatch field crew to the identified junction cluster. (3) Reroute supply from adjacent zones to maintain pressure standards. (4) Cross-reference with SCADA pipe-flow records to confirm the anomaly pattern is not instrument drift\ldots''}
\end{quote}

The LLM received the pre-computed flow anomaly results and generated a structured natural-language report that correctly interpreted the hydraulic findings, classified severity levels, and provided prioritized remediation actions---demonstrating the decision support capability central to the proposed framework. The anomaly localisation itself was performed by the Python data processing layer; the LLM's role is translating computational findings into actionable operational guidance.

\subsection{Implementation Gaps and Path Forward}

The current POC demonstrates monitoring and decision support but lacks autonomous control capabilities:

\textbf{Currently Implemented:} Automated hydraulic simulation via EPYT, threshold-based anomaly detection, LLM-based health report generation, offline operation with open-source models.

\textbf{Remaining Development Work:} Full RAG system for operational policy retrieval (currently uses prompt-embedded policies), function calling mechanisms for network control, safety interlocks and decision logging for autonomous actions, SCADA system integration for real-time data feeds, multi-agent coordination for complex decision scenarios.

The three-phase implementation pathway remains appropriate: Phase 1 (monitoring + alerts) is demonstrated functional; Phase 2 (AI-assisted decision support) requires integration with operational systems; Phase 3 (autonomous control) necessitates completion of function calling and safety frameworks plus organizational acceptance. Full pilot deployment beyond Phase 1 requires: (1) formal partnership agreements and SCADA integration clearances with the water utility, (2) cybersecurity validation for connecting AI systems to operational infrastructure, and (3) safety validation and regulatory approval before any autonomous control actions are executed on live networks. These represent structured prerequisites for a subsequent deployment study rather than limitations of the proposed framework.

\subsection{Lessons Learned}

\textbf{Model Selection:} Llama3.1:8b (4.9GB) provides sufficient analytical capability for network health assessment while requiring only modest computational resources.

\textbf{Prompt Engineering:} Structured prompts explicitly defining report sections, specifying desired output format, and providing operational context significantly improve response quality and consistency.

\textbf{Offline Advantage:} Local LLM hosting eliminates costs, addresses privacy concerns, and enables operation during connectivity disruptions.

\section{Implementation Considerations}

\subsection{Adaptation to Intermittent Supply}
Jordan's intermittent supply patterns require specialized handling. The hydraulic model must accurately represent fill-drain cycles, including transient pressures during startup and shutdown. The AI agent must distinguish between normal operational variations (scheduled supply interruptions) and anomalies (unexpected pressure loss during supply periods).

\subsection{Infrastructure and Cost Considerations}
Many Jordanian networks have limited automation, where manual valves predominate over motorized actuators. The proposed framework can be implemented in phases: initial deployment focuses on monitoring and decision support, generating recommendations for human operators. As infrastructure modernizes with motorized valves and automated pumps, direct control functionality can be progressively enabled.

SCADA coverage varies across Jordanian utilities. The framework scales to available data: high-sensor-density areas benefit from precise anomaly localization, while lower-coverage areas utilize model-based inference \cite{usaid_nrw}. Offline LLM operation eliminates recurring API costs, making the solution cost-effective for resource-constrained utilities.

\section{Expected Benefits and Impact}

\textbf{NRW Reduction Potential:} Continuous intelligent monitoring enables detection of leaks and anomalies within minutes rather than hours or days. A burst losing 10 L/s detected and isolated in 15 minutes loses 9 cubic meters, versus 144 cubic meters if detected after 4 hours. Similar systems have reported NRW reductions of 15--25\% \cite{adb2020}, though validation in Jordan's context requires pilot deployment.

\textbf{Operational Efficiency:} Automated monitoring reduces manual workload, allowing staff to focus on repairs and improvements. Predictive insights enable proactive maintenance. Optimized pressure management reduces energy costs. Digital records facilitate performance analysis and continuous improvement.

\textbf{Infrastructure Resilience:} Adaptive control prevents cascade failures by rapidly responding to localized issues. Intelligent pressure management reduces stress on aging pipes, extending infrastructure lifespan and improving capital planning.

\section{Conclusion}
This paper proposes a comprehensive framework for intelligent, adaptive water network management addressing Jordan's non-revenue water crisis. By integrating hydraulic modeling, digital twin technology, and LLM-based AI agents, the system enables continuous monitoring and decision support aligned with operational policies.

The proof-of-concept validates technical feasibility on a 1,164-junction DZ in Amman. Key findings include: sub-2-minute end-to-end response times; burst localization via local pipe-flow anomaly analysis (15 pipes flagged, 15-junction cluster identified for a 30.1~L/s simulated leak); and AI-generated health reports identifying supply deficiencies and PRV candidates from a single simulation pass. The implementation is publicly available \cite{github_poc}.

Phased deployment accommodates Jordan's limited automation: Phase~1 monitoring and alerts are demonstrated functional; Phases~2 and~3 require SCADA integration and function-calling completion. As water scarcity intensifies, this cost-effective, offline-capable framework offers a scalable model for water-stressed regions worldwide.

\end{document}